\documentclass[letterpaper]{article} 
\usepackage{aaai25}  
\usepackage{times}  
\usepackage{helvet}  
\usepackage{courier}  
\usepackage[hyphens]{url}  
\usepackage{graphicx} 
\usepackage{natbib}  
\usepackage{caption} 
\usepackage{algorithm}
\usepackage{algorithmic}
\usepackage{amsmath}
\usepackage{tabularx}
\usepackage{makecell}
\usepackage{multirow}
\usepackage{multicol}
\usepackage{booktabs}

\usepackage{newfloat}
\usepackage{listings}
\DeclareCaptionStyle{ruled}{labelfont=normalfont,labelsep=colon,strut=off} 
\floatstyle{ruled}
\newfloat{listing}{tb}{lst}{}
\floatname{listing}{Listing}
\nocopyright 

\title{Rethinking Relation Extraction: Beyond Shortcuts to Generalization with a Debiased Benchmark}
\author{
    Liang He, Yougang Chu, Zhen Wu, Jianbing Zhang, Xinyu Dai, Jiajun Chen \\
}
\affiliations{
    National Key Laboratory for Novel Software Technology, Nanjing University

    heliang@smail.nju.edu.cn
}

\usepackage{bibentry}

\begin{document}

\maketitle

\begin{abstract}
Benchmarks are crucial for evaluating machine learning algorithm performance, facilitating comparison and identifying superior solutions. However, biases within datasets can lead models to learn shortcut patterns, resulting in inaccurate assessments and hindering real-world applicability. This paper addresses the issue of entity bias in relation extraction tasks, where models tend to rely on entity mentions rather than context. We propose a debiased relation extraction benchmark DREB that breaks the pseudo-correlation between entity mentions and relation types through entity replacement. DREB utilizes Bias Evaluator and PPL Evaluator to ensure low bias and high naturalness, providing a reliable and accurate assessment of model generalization in entity bias scenarios. To establish a new baseline on DREB, we introduce MixDebias, a debiasing method combining data-level and model training-level techniques. MixDebias effectively improves model performance on DREB while maintaining performance on the original dataset. Extensive experiments demonstrate the effectiveness and robustness of MixDebias compared to existing methods, highlighting its potential for improving the generalization ability of relation extraction models. We will release DREB and MixDebias publicly.
\end{abstract}

%

\section{Introduction}


Benchmarks are crucial for evaluating machine learning algorithms, providing standardized datasets to compare methods and identify top performers. However, reliance on specific datasets can introduce biases, causing models to learn shortcut patterns instead of true semantic understanding, which hinders their real-world applicability. Studies show that improved performance often stems from exploiting dataset biases rather than enhanced comprehension. For example, in natural language inference, models tend to predict based on lexical overlap ratios or the presence of negation words on SNLI \cite{bowman2015large} and MNLI \cite{williams2018broad} datasets \cite{gururangan2018annotation,mccoy2019right}, and in fact verification tasks, they often rely on specific phrases rather than the contextual relationship between claims and evidence \cite{schuster2019towards}.

\begin{figure}[htbp]
    \centering
    \includegraphics[width=0.4\textwidth]{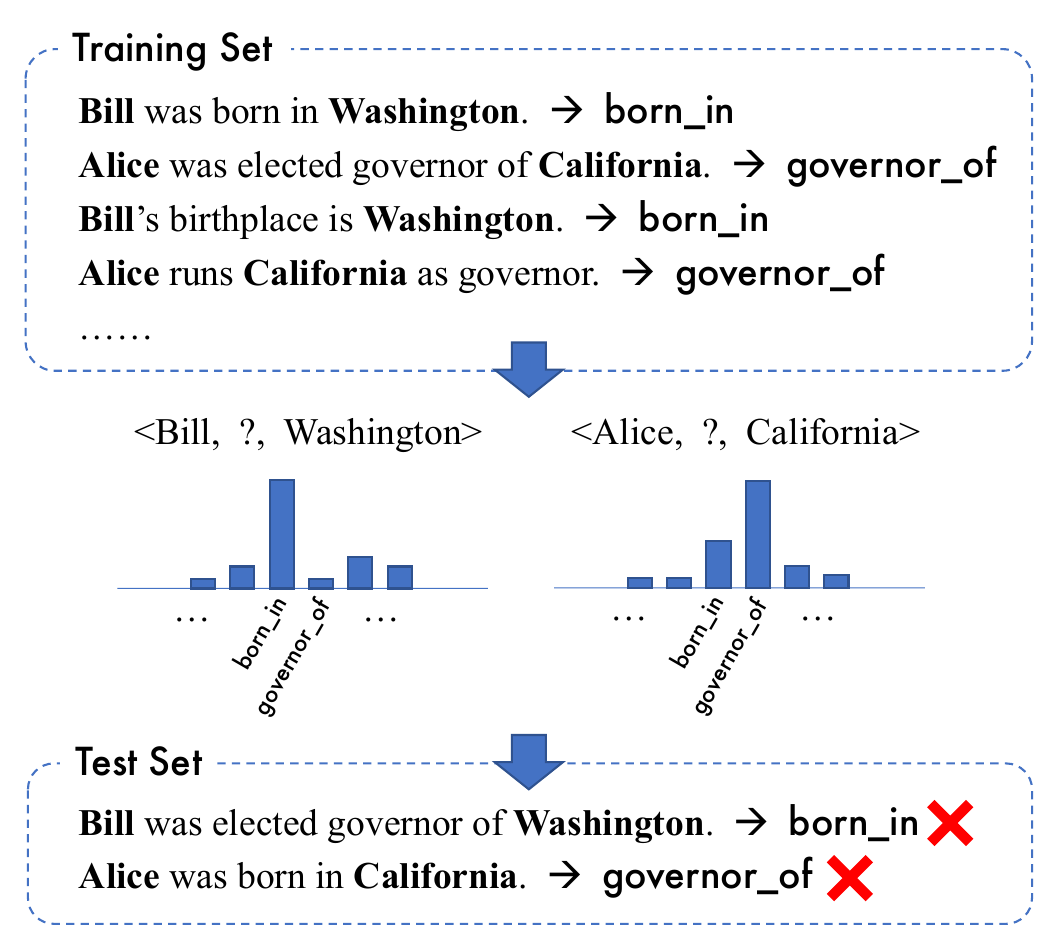}
    \caption{An illustrative example of how entity biases can cause models to learn false shortcuts, inevitably resulting in erroneous predictions.}
    \label{fig:entity-bias-example}
\end{figure}


In relation extraction tasks, widely-used datasets like SemEval 2010 Task 8 \cite{hendrickx2010semeval}, TACRED \cite{zhang2017position}, TACREV \cite{alt2020tacred}, and Re-TACRED \cite{stoica2021re} exhibit entity bias, where entity mentions can provide superficial cues for relation types (Figure \ref{fig:entity-bias-example}). This pseudo-correlation between entity mentions and relation types means models can often predict accurately without textual context \cite{zhang2018graph,peng2020learning}. For instance, over half of TACRED instances can be correctly predicted using only entity mentions \cite{wang2022should}. After entity replacement, state-of-the-art models like LUKE \cite{yamada2020luke} and IRE \cite{zhou2022improved} experience significant drops in performance (30\% - 50\% F1 score) \cite{wang2023fragile}. Large language models exacerbate this bias by disregarding contradictory or underrepresented contextual information, overly relying on biased parametric knowledge \cite{longpre2021entity} for predictions \cite{wang2023causal}. These findings highlight a critical over-reliance on entity mentions, severely impacting model performance when entity mentions are absent or debiased.


To address the entity bias issue, various approaches have been explored at both the data and model levels. However, existing works still face challenges: At the data level, debiasing methods may inadvertently introduce new biases, compromising evaluation reliability. For instance, \cite{wang2022should}'s modification of the TACRED and Re-TACRED datasets results in distribution bias due to changes in relation type distribution, and ENTRED's entity replacement \cite{wang2023fragile} lacks semantic constraints, potentially introducing semantic bias. At the model level, DFL \cite{mahabadi2020end} modifies the focal loss function to reduce focus on biased samples but may damage in-domain performance and the learning of useful features while reducing bias. R-Drop \cite{liang2021r} uses regularization to decrease reliance on specific features but lacks fine-grained control over entity biases. CoRE \cite{wang2022should}, employing counterfactual analysis, may not fully mitigate biases learned during training due to its post-processing nature.


To evaluate the generalization of relation extraction models under entity bias, we design a debiased benchmark DREB using entity replacement to break the pseudo-correlation between entity mentions and relation types. We employed Bias Evaluator and PPL Evaluator to ensure low bias and high naturalness of the benchmark. To establish a baseline on DREB, we proposed MixDebias, a method that combines data-level augmentation with model-level debiasing. At the data level, it generates augmented samples and uses Kullback-Leibler (KL) divergence \cite{belov2011distributions} to align probability distributions. At the model level, a bias model assesses sample bias, and a debiased loss function optimizes the model. Experiments show that MixDebias significantly enhances model performance on DREB while maintaining stability on the original dataset.

Our contribution can be summarized as three-fold:
\begin{itemize}
    \item Firstly, we propose a debiased relation extraction benchmark DREB that ensures models cannot rely solely on entity mentions for prediction. Using the Bias Evaluator and PPL Evaluator, DREB offers low bias and high naturalness, providing a more reliable assessment dataset for measuring model generalization in entity bias scenarios.
    \item Secondly, we introduce MixDebias, a new baseline that enhances model performance on DREB through combined debiasing at the data and model training levels while maintaining performance on the original dataset.
    \item Finally, we conduct a comprehensive evaluation and comparison of existing relation extraction models and debiasing methods. Our experiments show that DREB can better evaluate the debiasing capability of relation extraction models, and MixDebias achieves excellent performance across multiple datasets, verifying its effectiveness and robustness.
\end{itemize}

\section{Related Work}

For debiasing in relation extraction, efforts have focused on both data and model levels. \textbf{Data Level}: \cite{wang2022should} introduces a filtered evaluation setting based on the TACRED dataset, retaining only samples where the relation cannot be accurately predicted using just the entity pairs. ENTRED \cite{wang2023fragile} employs type-constrained and random entity replacements to assess model robustness. Type-constrained replacement maintains entity class consistency, while random replacement introduces diversity. \textbf{Model Level}: DFL \cite{mahabadi2020end} adjusts the loss function based on bias-only model predictions, enabling the model to focus more on hard examples and less on biased ones. R-Drop \cite{liang2021r} enforces consistency among output distributions of sub-models generated by dropout, improving generalization. CoRE \cite{wang2022should} constructs a causal graph to identify and mitigate biases caused by reliance on entity mentions, focusing predictions more on textual context.

\section{DREB: A Debiased Relation Extraction Benchmark}

\begin{figure*}[ht]
    \centering
    \includegraphics[width=\linewidth]{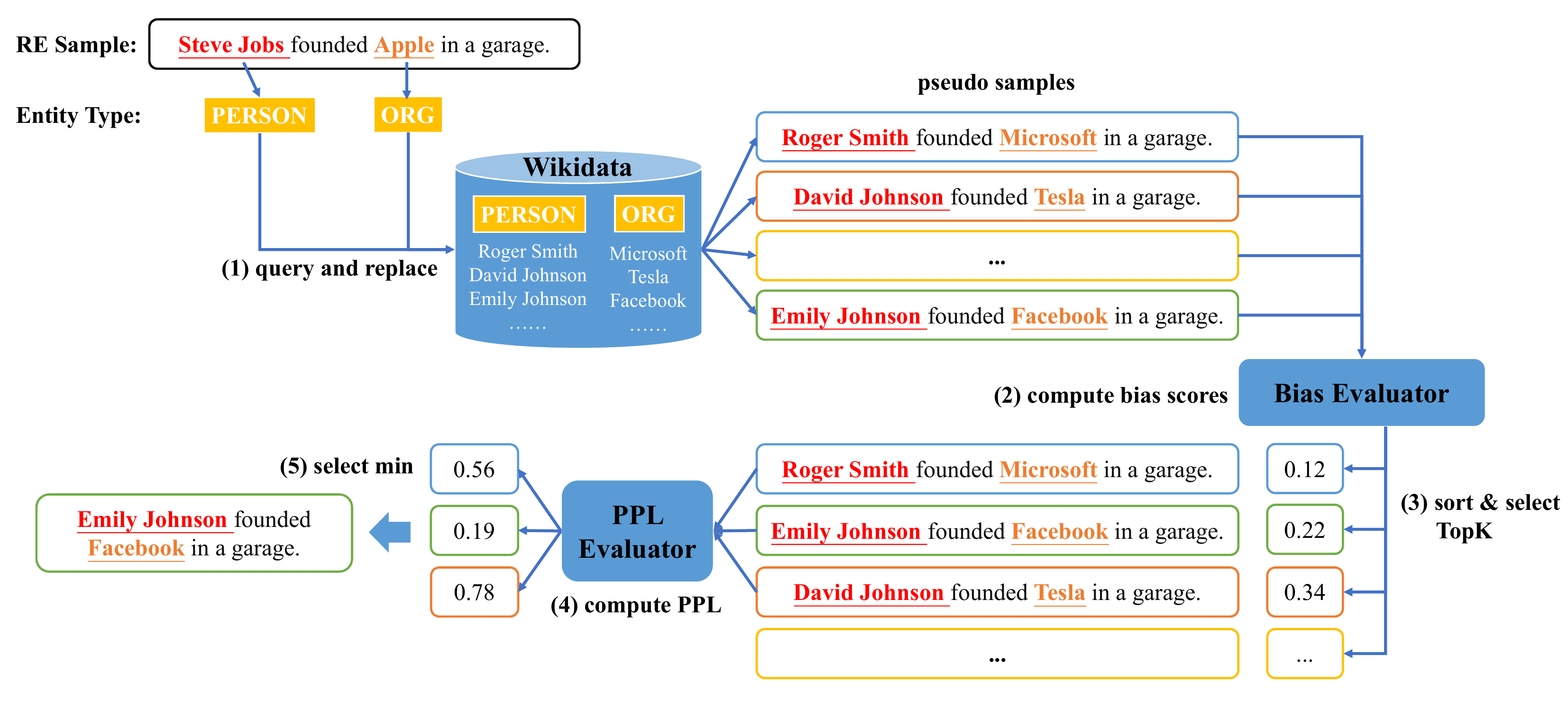}
    \caption{The construction workflow of DREB benchmark.}
    \label{fig:workflow}
\end{figure*}


We introduce DREB, a debiased relation extraction benchmark designed to dismantle pseudo-correlations between entity mentions and relation types, preventing models from solely inferring relations based on entity mentions. As illustrated in Figure \ref{fig:workflow}, DREB construction involves substituting entities in the test set with entities of the same type from Wikidata \cite{vrandevcic2014wikidata} to generate pseudo samples. Our method uniquely incorporates a Bias Evaluator to select replacements with minimal bias and a PPL Evaluator to ensure the naturalness and quality of the pseudo samples.

\subsubsection{Bias Evaluator.}
Bias is fundamentally a pseudo-correlation between biased dataset features and their corresponding labels. To counter this, we employ a neural network to model these correlations directly. Given a sample denoted by \( x \) and its corresponding label \( y \), the process of extracting bias features from \( x \) is represented by \( \phi(x) \). By training the network \( F : \phi(x) \rightarrow y \), the output \( F(\phi(x)) \) reflects the bias inherent in \( x \). For entity bias specifically, the feature extraction process \( \phi \) is defined such that it captures the essence of the entity bias present in relation extraction samples. For instance, \( \phi(\text{"Steve Jobs founded Apple in a garage."}) \) would yield "Steve Jobs" and "Apple." We preprocess the relation extraction training set \( \mathcal{D} \) with \( \phi(x) \) to construct a synthetic dataset \( \mathcal{D}_{\text{EntityBias}} \), which allows us to model the entity bias directly. The resulting model, once trained, serves as a bias evaluator to measure the degree of entity bias in pseudo samples.

\subsubsection{PPL Evaluator.}
Entity replacement schemes generate synthetic text data, which may result in some degree of unnaturalness. To improve the quality of the challenge set, we generate multiple synthetic samples in batches and use GPT-2 \cite{radford2019language} as a language model to calculate the perplexity of these samples. Given a sequence \( \mathbf{W} = (w_1, w_2, \ldots, w_n) \), where \( w_i \) is the \( i \)-th word and \( n \) is the number of words in the sequence, the perplexity can be calculated using the following formula:

\begin{equation}
    \begin{aligned}
        \log \text{PPL}(W) &= \log \left( \frac{1}{P(w_1, w_2, \ldots, w_n)} \right)^{\frac{1}{n}} \\
        &= -\frac{1}{n} \sum_{i=1}^{n} \log P(w_i|w_1, \ldots, w_{i-1})
    \end{aligned}
\end{equation}

\noindent where \( P(w_1, w_2, \ldots, w_n) \) is the probability of the sequence \( \mathbf{W} \). We then select the sample with the lowest perplexity as the final generated sample. Through this process, we can filter out the most natural samples according to the language model, thus enhancing the naturalness and overall quality of the challenge set.

We selected widely used relation extraction datasets TACRED, TACREV, and Re-TACRED and applied our proposed debiasing dataset construction strategy to build DREB benchmark. These datasets belong to the sentence-level relation extraction category, where TACRED is the initial version, TACREV is a revised version that addresses annotation and noise issues in the test and validation sets of TACRED, and Re-TACRED redesigns the relation types and the dataset itself.

\subsection{Benchmark Analysis}

\subsubsection{Does DREB introduce distribution biases?}
\begin{figure}[ht]
    \centering
    \includegraphics[width=\linewidth]{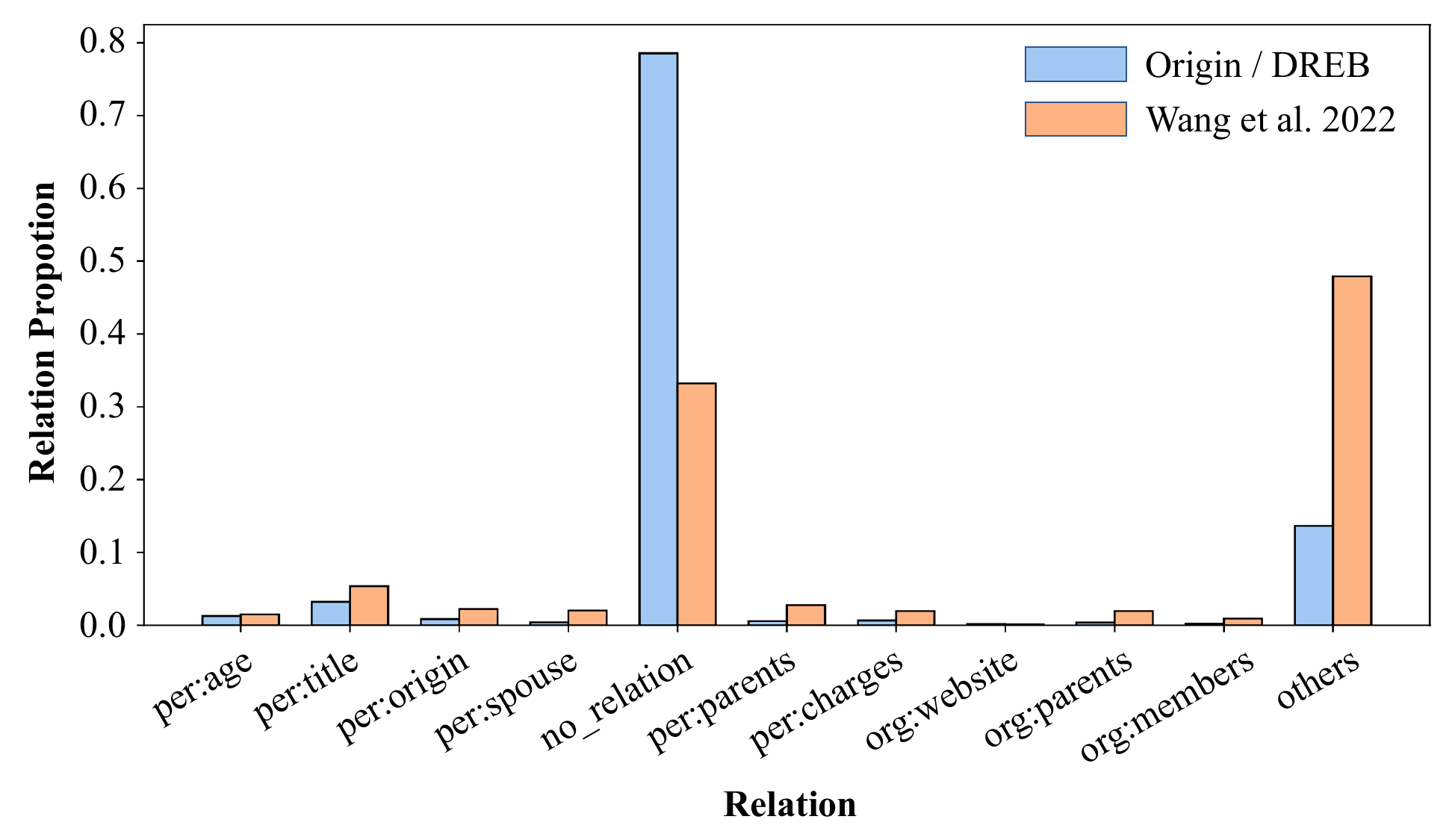}
    \caption{Comparison of relation type distributions.}
    \label{fig:distribution_bias}
\end{figure}

Figure \ref{fig:distribution_bias} compares the relation distributions between DREB, the original datasets, and the method by \cite{wang2022should}. It shows that the datasets constructed by \cite{wang2022should}'s method exhibit significant shifts in relation distribution compared to the original datasets, particularly with a notable reduction in the proportion of \textbf{no\_relation}. This suggests that models could simply lower their classification thresholds to boost recall, artificially inflating evaluation metrics. In contrast, DREB maintains identical relation distributions to the original datasets, avoiding the introduction of new distribution biases and ensuring the accurate assessment of debiasing methods.

\subsubsection{Does DREB introduce semantic biases?}
\begin{figure}[ht]
    \centering
    \includegraphics[width=\linewidth]{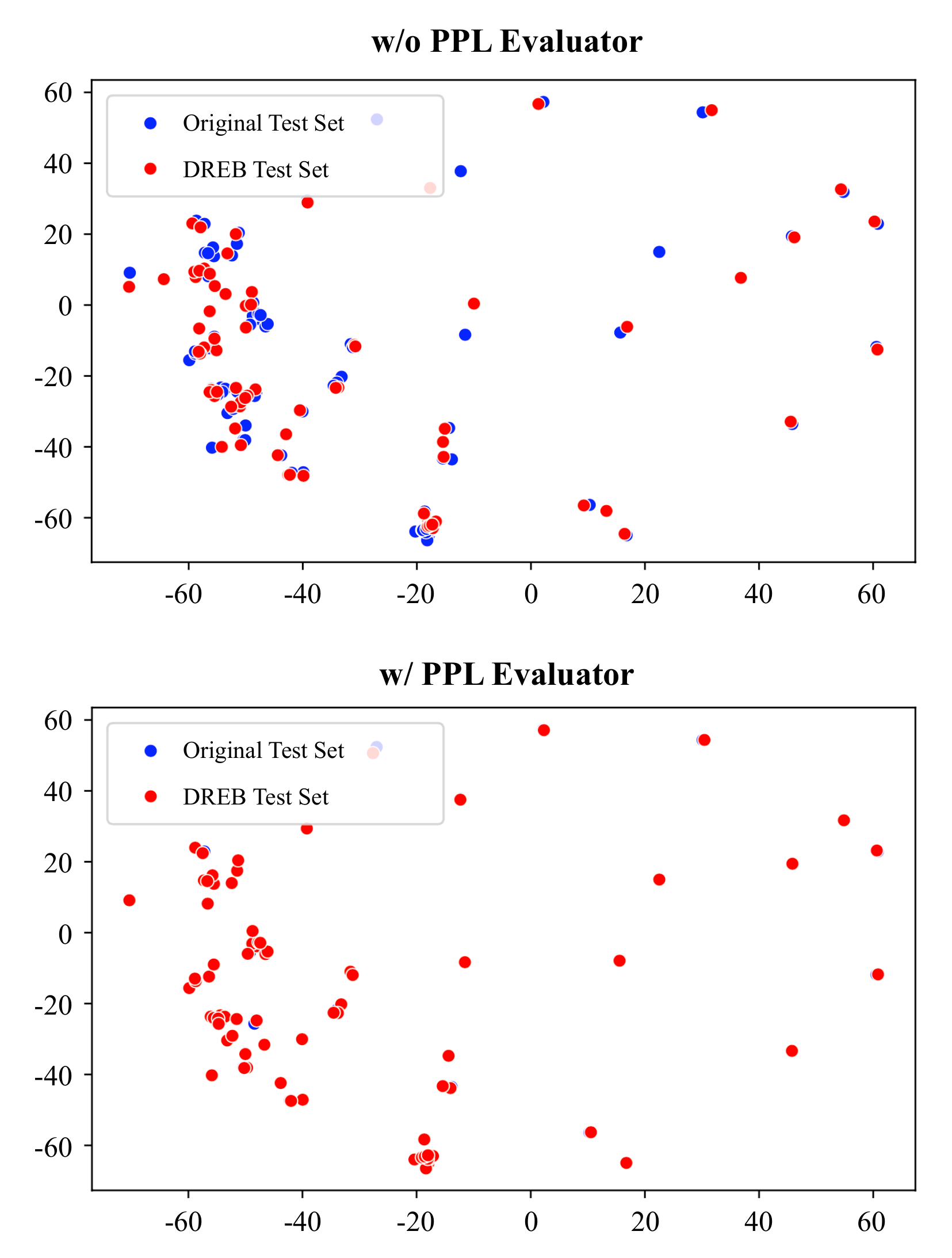}
    \caption{Comparison of semantic distributions. The PPL Evaluator can effectively control semantic bias.}
    \label{fig:semantic_bias}
\end{figure}

We compared the semantic distribution differences between DREB test set samples generated with and without the PPL Evaluator (w/ and w/o PPL Evaluator, respectively) and the original test set samples. As shown in Figure \ref{fig:semantic_bias}, we used SBERT \cite{reimers2019sentence} to encode the samples into feature vectors and then applied PCA \cite{smith2002tutorial} to reduce them to a 2D space for visualization. Without the PPL Evaluator, there was a noticeable distribution shift in the generated samples compared to the original samples, introducing semantic bias. However, when the PPL Evaluator was used, the generated samples largely overlapped with the original samples, avoiding the introduction of semantic bias. This visualization demonstrates that the PPL Evaluator effectively maintains the continuity of samples in the semantic space during the generation of DREB samples, ensuring their semantic naturalness and consistency with the original dataset samples.

\section{MixDebias: A New Baseline on DREB}

\begin{figure*}[ht]
    \centering
    \includegraphics[width=\linewidth]{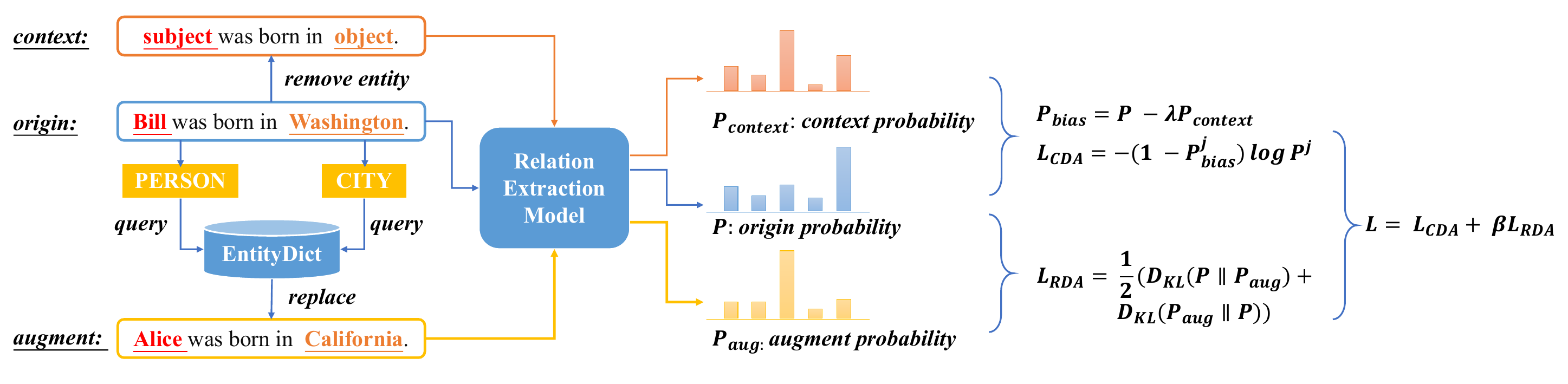}
    \caption{The overall workflow of MixDebias.}
    \label{fig:mixdebias}
\end{figure*}

Based on the DREB, we also introduce a method called MixDebias as a new baseline, which debiases from both the data and model training levels (Figure \ref{fig:mixdebias}).

\subsubsection{Data-level debiasing (RDA, Regularized Debias Approach):} 
Entity mentions, despite their potential to cause bias, are valuable as they can prevent ambiguity, particularly in sentences with multiple entities of the same type. Instead of simplistically substituting entities with their corresponding entity types, we propose an approach that generates multiple data-augmented samples from an original training sample through entity replacement. This process is guided by a Kullback-Leibler Divergence (KL Divergence) constraint that encourages the model to produce probability distributions \( P \) and \( P_{aug} \) that are as similar as possible when presented with the original and augmented samples, respectively. We term this KL divergence constraint \( \mathcal{L}_{RDA} \), and its incorporation effectively reduces the model's reliance on the entities present in the input, thereby enhancing the model's generalization capabilities.

Specifically, we construct an entity dictionary (EntityDict) by extracting entities from the training set, facilitating the dynamic creation of data-augmented samples during training through entity replacement. We deliberately avoid sourcing entities from external resources like Wikidata for augmentation to prevent the introduction of lexical bias during the training phase. Throughout training, for an original sample, we dynamically retrieve entities of the same type from the EntityDict and generate a new data-augmented sample via entity replacement. Both the original and augmented samples are then fed into the relation extraction model, yielding two probability distributions, \( P \) and \( P_{aug} \). We calculate the KL divergence between these distributions. Due to the asymmetry of KL divergence, we calculate \(D_{KL}(P || P_{aug})\) and \(D_{KL}(P_{aug} || P)\) and average them to get \(\mathcal{L}_{RDA}\).

\subsubsection{Model-level debiasing (CDA, Casual Debias Approach):}
The CDA method identifies and quantifies entity bias through causal effect estimation and uses this estimation to guide model training, reducing the model's dependence on input features that may lead to bias. In causal effect estimation, we try to understand how different factors affect the results, especially how other variables affect the results when some variables are controlled. For relation extraction models, causal effects can be used to identify and reduce the model's dependence on input features that may have pseudo-correlation with the target output, rather than real causal relationships. The CDA method uses causal effect estimation to build a bias model (Bias Model), which assesses the degree of entity bias in each sample. Specifically, by providing only the context input to the model to obtain the probability distribution \( P_{context} \), and the original sample input to the model to obtain the probability distribution \( P \), then calculate \( P - \lambda P_{context} \) to obtain the bias probability distribution \( P_{bias} \), where \( \lambda \) is a hyperparameter. This bias probability distribution reflects the degree of entity bias in the sample. The CDA method uses Debiased Focal Loss \cite{mahabadi2020end} for model training, which adjusts the model's predictions using the bias probability, thereby reducing the model's dependence on entity mentions.

\begin{equation}
    \mathcal{L}_{CDA} = -(1 - P_{bias}^j) \log P^j
\end{equation}

\noindent where \( j \) is the correct relation type label. When \( \lambda \) is 0, \( \mathcal{L}_{CDA} \) degenerates into \( -(1 - P^j) \log P^j \), which is the Focal Loss. As a common form of model regularization loss, we modify it with \( P_{context} \) to achieve a debiasing effect. In this way, the CDA method reduces the entity bias learned by the model during the training process, improving the model's generalization ability when facing different entities.

Finally, we introduce a hyperparameter \( \beta \) to combine \( \mathcal{L}_{RDA} \) and \( \mathcal{L}_{CDA} \) in a weighted manner to obtain the final loss function \( \mathcal{L}_{MixDebias} \):

\begin{equation}
    \begin{aligned}
        \mathcal{L}_{MixDebias} &= \mathcal{L}_{CDA} + \beta \mathcal{L}_{RDA} \\
        &= -(1 - P_{bias}^j) \log P^j + \\
        & \quad \frac{\beta}{2} (D_{KL} (P || P_{aug}) + D_{KL} (P_{aug} || P)) \\
        &= - (1 - (P^j - \lambda P_{context}^j)) \log P^j + \\
        & \quad \frac{\beta}{2} (D_{KL} (P || P_{aug}) + D_{KL} (P_{aug} || P))
    \end{aligned}
\end{equation}

\subsection{Evaluation}

\subsubsection{Evaluation metric.}
Consistent with previous work, we adopt the F1-score, which is the harmonic mean of precision and recall, as our primary evaluation metric. Additionally, we designed the Bias Mitigation Efficiency (BME) to comprehensively evaluate the effectiveness of debiasing methods, taking into account both the performance impact on the original dataset and the performance improvement on DREB. Specifically, let $\widetilde{F1}_{\text{origin}}$ and \( \widetilde{F1}_{\text{DREB}} \) be the F1 scores of the baseline model on the original dataset and DREB, respectively. Let \( F1_{\text{origin}} \) and \( F1_{\text{DREB}} \) be the F1 scores of the new model on the original dataset and DREB, respectively. The BME is then calculated as:

\begin{equation}
    \text{BME} = \alpha \cdot \frac{F1_{\text{origin}}}{\widetilde{F1}_{\text{origin}}} + (1 - \alpha) \cdot \frac{F1_{\text{DREB}}}{\widetilde{F1}_{\text{DREB}}}
\end{equation}

\noindent where in our experiments, we set \( \alpha = 0.5 \).

\subsubsection{Baselines.} 
To focus on analyzing the debiasing effects of the model, models that retain entity mentions in the input during the preprocessing stage better meet our needs. We selected \textbf{LUKE} \cite{yamada2020luke} and \textbf{IRE} \cite{zhou2022improved} for this purpose. LUKE is a transformer-based model that introduces a novel pretraining task for learning contextualized representations of both words and entities. IRE introduces typed entity markers that include both the entity spans and their types into the input text, allowing for a more comprehensive representation of entity mentions. In terms of debiasing methods, we primarily chose the following as baseline methods for comparison: \textbf{Focal} \cite{lin2018focal} reduces the model's reliance on entities by attenuating the influence of easily classified samples and amplifying the significance of challenging ones, thereby recalibrating the training focus towards hard-to-classify instances. \textbf{R-Drop} \cite{liang2021r} enhances model generalization by enforcing consistency between output distributions of sub-models created through dropout, processing each mini-batch data sample twice to generate distinct outputs, and minimizing the bidirectional Kullback-Leibler divergence, thereby reducing reliance on entity mentions and improving the model's robustness. \textbf{DFL} \cite{mahabadi2020end} adjusts the loss function using a focusing parameter based on the bias-only model's predictions, effectively reducing the model's dependency on entities by downweighting samples with high entity bias, which enhances the model's robustness and generalization without altering its original architecture. \textbf{PoE} \cite{hinton2002training} employs a unique integration of individual expert models by multiplying their probability distributions, including a biased distribution derived solely from entity inputs, with the model's predictive distribution. This multiplication and subsequent renormalization subtly decrease the influence of samples with significant entity bias, effectively reducing the model's reliance on these entities while optimizing model performance. \textbf{CoRE} \cite{wang2022should} mitigates biases by constructing a causal graph to identify dependencies and using counterfactual scenarios to pinpoint entity biases, subsequently refining predictions through an adaptive bias mitigation process that emphasizes textual context over entity reliance, leading to debiased outcomes.

\subsubsection{Main results.}

Table \ref{table:main} demonstrates the performance comparison of various relation extraction models and different debiasing methods on different datasets, where \( F1_{\text{origin}} \) represents the performance on the original test set, and \( F1_{\text{DREB}} \) represents the performance on DREB benchmark proposed in this paper.

\begin{table*}[ht]
\centering
    \begin{tabularx}{\linewidth}{p{2.9cm}|p{1.2cm}<{\centering}p{1.2cm}<{\centering}p{1.2cm}<{\centering}|p{1.2cm}<{\centering}p{1.2cm}<{\centering}p{1.2cm}<{\centering}|p{1.2cm}<{\centering}p{1.2cm}<{\centering}p{1.2cm}<{\centering}}
    \toprule

    \multirow{2}*{\textbf{Model}} & \multicolumn{3}{c|}{\textbf{TACRED}} & \multicolumn{3}{c|}{\textbf{TACREV}} & \multicolumn{3}{c}{\textbf{Re-TACRED}} \\
    \cmidrule(lr){2-10}
    & $F1_{\text{origin}}$ & $F1_{\text{DREB}}$  & BME & $F1_{\text{origin}}$  & $F1_{\text{DREB}}$  & BME & $F1_{\text{origin}}$  & $F1_{\text{DREB}}$  & BME \\
   
    \midrule
        LUKE & 70.82 & 44.40 & - & 80.16 & 50.60 & - & 88.92 & 39.40 & - \\
        +Focal & 69.94 & 45.55 & 1.01 & 79.15 & 52.48 & 1.01 & 88.58 & 39.32 & 1.00 \\
        +R-Drop & \textbf{70.99} & 46.68 & 1.03 & \textbf{81.06} & 53.85 & 1.04 & \textbf{89.53} & 40.89 & 1.02 \\
        +DFL & 65.04 & 48.48 & 1.01 & 71.31 & 53.17 & 0.97 & 84.15 & 43.94 & 1.03 \\
        +PoE & 63.32 & 47.63 & 0.98 & 68.82 & 52.02 & 0.94 & 82.46 & 44.10 & 1.02 \\
        +CoRE & 70.04 & 47.87 & 1.03 & 79.82 & 54.88 & 1.04 & 87.13 & 41.94 & 1.02 \\
        +MixDebias & 69.93 & \textbf{62.44} & \textbf{1.20} & 80.91 & \textbf{72.93} & \textbf{1.23} & 87.95 & \textbf{77.71} & \textbf{1.48} \\
    \midrule
        IRE & 71.27 & 50.94 & - & 79.36 & 57.20 & - & 87.43 & 46.25 & - \\
        +Focal & 71.11 & 50.97 & 1.00 & 78.55 & 57.51 & 1.00 & 87.51 & 48.22 & 1.02 \\
        +R-Drop & 71.13 & 52.98 & 1.02 & 80.37 & 59.71 & 1.03 & \textbf{87.96} & 48.40 & 1.03 \\
        +DFL & 65.72 & 56.28 & 1.01 & 70.18 & 60.13 & 0.97 & 80.17 & 54.03 & 1.04 \\
        +PoE & 64.72 & 54.67 & 0.99 & 69.12 & 59.28 & 0.95 & 81.35 & 51.41 & 1.02 \\
        +CoRE & 70.43 & 55.00 & 1.03 & 78.82 & 60.81 & 1.03 & 86.21 & 48.36 & 1.02 \\
        +MixDebias & \textbf{71.99} & \textbf{70.02} & \textbf{1.19} & \textbf{80.97} & \textbf{79.15} & \textbf{1.20} & 87.27 & \textbf{82.17} & \textbf{1.39} \\
    \bottomrule
    \end{tabularx}
\caption{The overall evaluation results. MixDebias significantly enhances performance on DREB and achieves comparable performance to the best models on the original dataset. In terms of the comprehensive metric BME, MixDebias also leads far ahead of other baseline methods.}
\label{table:main}
\end{table*}

\begin{table*}[ht]
\centering
    \begin{tabularx}{\linewidth}{p{2.95cm}|p{2cm}<{\centering}p{2cm}<{\centering}|p{2cm}<{\centering}p{2cm}<{\centering}|p{2cm}<{\centering}p{2cm}<{\centering}}
    \toprule

    \multirow{2}*{\textbf{Model}} & \multicolumn{2}{c|}{\textbf{TACRED}} & \multicolumn{2}{c|}{\textbf{TACREV}} & \multicolumn{2}{c}{\textbf{Re-TACRED}} \\
    \cmidrule(lr){2-7}
    & $F1_{\text{origin}}$  & $F1_{\text{DREB}}$ & $F1_{\text{origin}}$  & $F1_{\text{DREB}}$ & $F1_{\text{origin}}$  & $F1_{\text{DREB}}$ \\
   
    \midrule
        LUKE+MixDebias & 69.93 & 62.44 & 80.91 & 72.93 & 87.95 & 77.71 \\
        -CDA & 69.66(-0.27) & 62.06(-0.38) & 80.63(-0.28) & 71.99(-0.94) & 88.45(+0.50) & 78.26(+0.55) \\
        -RDA & 69.68(-0.25) & 45.77(-16.67) & 79.32(-1.59) & 51.91(-21.02) & 88.69(+0.74) & 39.72(-37.99) \\
    \midrule
        IRE+MixDebias & 71.99 & 70.02 & 80.97 & 79.15 & 87.27 & 82.17 \\
        -CDA & 71.92(-0.07) & 70.21(+0.19) & 80.78(-0.19) & 78.60(-0.55) & 87.19(-0.08) & 82.08(-0.09) \\
        -RDA & 71.33(-0.66) & 52.60(-17.42) & 79.36(-1.61) & 58.48(-20.67) & 87.87(+0.60) & 48.22(-33.95) \\
    \bottomrule
    \end{tabularx}
\caption{The ablation study results for the MixDebias method, detailing the performance impacts of individual components CDA and RDA.}
\label{table:ablation}
\end{table*}


The experimental outcomes yield these insights: \textbf{LUKE and IRE} experienced a notable decline in performance on the DREB, suggesting their initial high results were partially due to reliance on entity mentions that were either removed or disguised in the DREB context, thereby affecting their efficacy. \textbf{Focal and R-Drop}, though not originally intended to tackle entity bias, have still been found to alleviate it. These techniques, primarily targeting overfitting, incidentally lessen the models' dependency on entity cues, indicating that generalization-focused strategies can also indirectly benefit bias reduction. \textbf{DFL and PoE}, as targeted debiasing approaches, markedly bolstered model performance on DREB through the incorporation of bias evaluation and adjustment within the training regime. However, this enhancement seems to have compromised the models' performance on the original data. \textbf{CoRE}, tailored to counteract entity bias, successfully improved DREB performance without sacrificing the original dataset's results, reflecting a balanced and potent debiasing approach. In sum, our proposed \textbf{MixDebias} method has impressively uplifted performance on DREB while also maintaining or even enhancing the original dataset's performance, showcasing its robust adaptability and debiasing capabilities.

\begin{figure*}[ht]
    \centering
    \includegraphics[width=\linewidth]{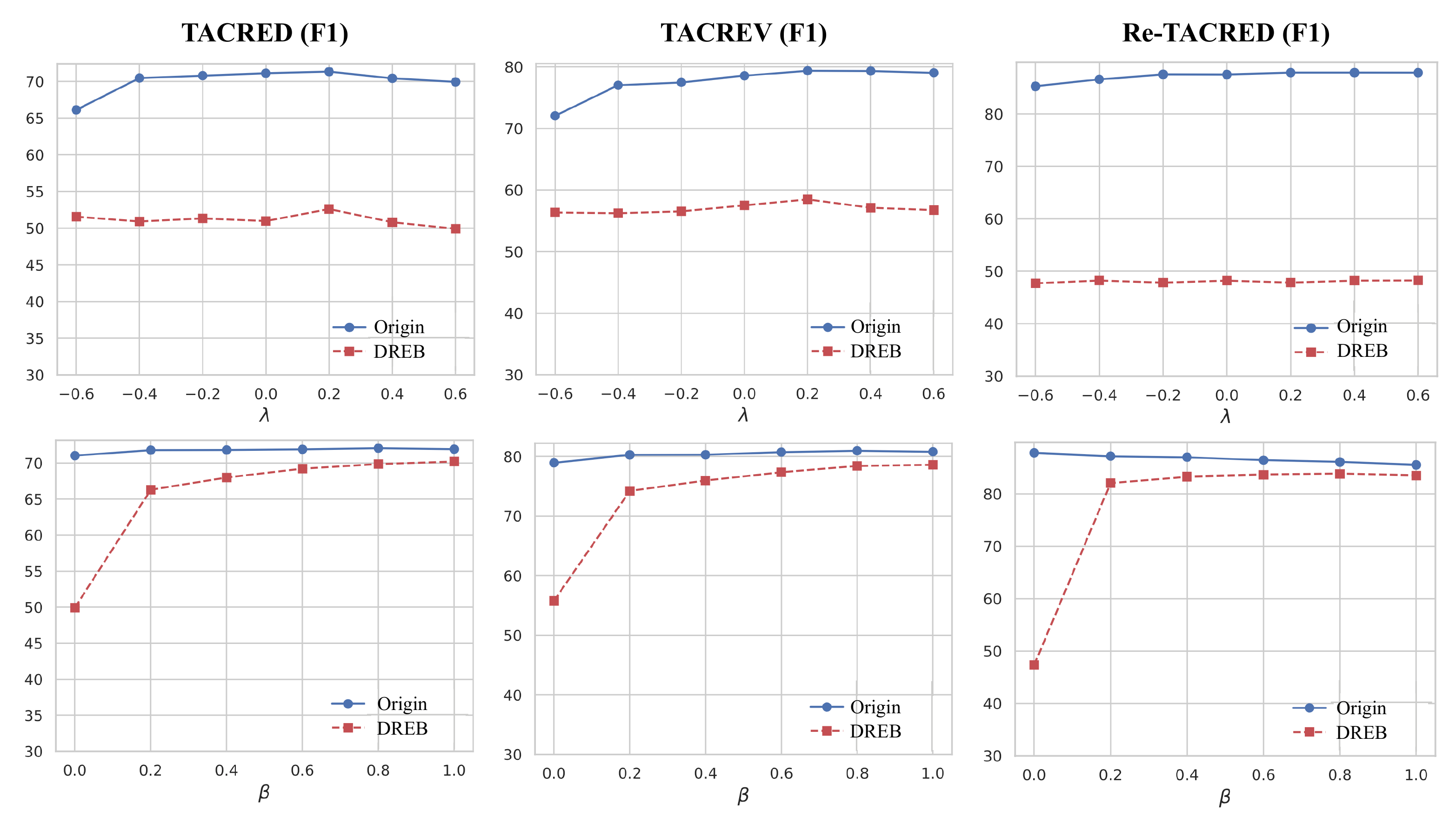}
    \caption{The detailed ablation analysis on the hyperparameters \( \lambda \) and \( \beta \) in MixDebias.}
    \label{fig:parameter}
\end{figure*}

\begin{figure*}[!]
    \centering
    \includegraphics[width=\linewidth]{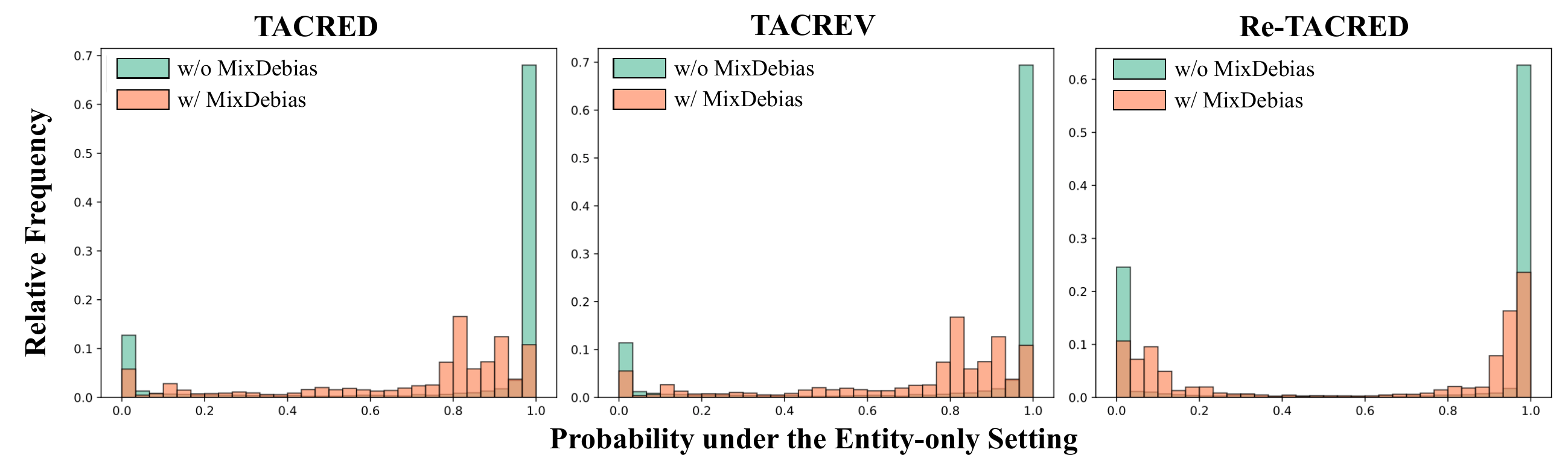}
    \caption{The visualization of debiasing effect. The substantial reduction in model reliance on entity mentions with MixDebias leads to a more uniform probability distribution.}
    \label{fig:generalization}
\end{figure*}

\subsubsection{Ablation study.}

As shown in Table \ref{table:ablation}, we conducted an ablation study on the two components of MixDebias, RDA and CDA. From the experimental results, we can draw the following conclusions: Both RDA and CDA are effective methods for removing entity bias. Overall, RDA is more effective than CDA. However, in most scenarios, these two methods are complementary and can enhance performance on DREB while minimizing the impact on the performance of the original dataset.


At the same time, we conducted a more detailed ablation analysis on the hyperparameters \( \beta \) and \( \lambda \) in MixDebias. Here, \( \beta \) represents the weight of the KL divergence, with a value range of [0.0, 1.0]; and \( \lambda \) represents the hyperparameter for estimating the biased probability distribution of samples using causal effects, with a value range of [-0.6, 0.6]. From Figure \ref{fig:parameter}, we can draw the following conclusions: When \( \beta = 0 \), it is equivalent to the model not considering RDA. However, when \(\beta \neq 0\), introducing RDA leads to significant performance improvements, and as \(\beta\) increases, the debiasing effect becomes stronger. Particularly on noisy datasets such as TACRED and TACREV, the model also shows a slight performance improvement on the original dataset. Compared to \(\beta\), the \(\lambda\) parameter has a smaller impact on model performance. When \(\lambda = 0.2\), the model performs optimally. This suggests that after applying the RDA method, the level of entity bias in the samples is already significantly reduced. In this case, CDA mainly addresses the bias that is difficult to correct at the data level, serving as a complementary effect to RDA, thereby further reducing the model's reliance on entities.

\subsubsection{Model generalization analysis.}

As shown in Figure \ref{fig:generalization}, we plotted the label probability distribution of the model under the setting of entity-only input in the original test set before and after debiasing on the TACRED, TACREV, and Re-TACRED datasets. The experimental results show that for the baseline relation extraction model, under the entity-only input setting, the label probabilities are primarily concentrated around values close to 1, indicating that entity mentions significantly influences the model's prediction outcomes. After applying MixDebias debiasing method, the output probabilities of the model become notably more uniform. At this point, the pseudo-correlation between entity mentions and relation types is significantly reduced, decreasing the likelihood of the model making incorrect predictions due to entity misguidance, thus enhancing the model's generalization capability.

\section{Conclusion}

This paper introduces DREB, a debiased relation extraction benchmark, and MixDebias, a novel debiasing method that addresses entity bias in relation extraction models.
DREB's strength lies in its ability to sever spurious links between entity mentions and relation types through strategic entity replacement, fostering a benchmark with diminished bias and elevated naturalness. This is achieved with Bias Evaluator and PPL Evaluator, which ensure the benchmark maintains a high standard of impartiality and linguistic authenticity. MixDebias enhances model performance on DREB while maintaining robustness on the original dataset through a combination of data-level augmentation and model-level debiasing strategies. Comprehensive experiments demonstrate MixDebias's effectiveness in improving model generalization and reducing reliance on entity mentions, setting a new standard for debiasing in relation extraction tasks.


\bibliography{ref}

\end{document}